\documentclass[a4paper]{article}
\usepackage{interspeech2013,amssymb,amsmath,epsfig,makecell,multirow}
\pdfoutput=1
\ninept
\def\reg{{\rm\ooalign{\hfil
     \raise.07ex\hbox{\scriptsize R}\hfil\crcr\mathhexbox20D}}}

\title{Feature Learning with Gaussian Restricted Boltzmann Machine for \\Robust Speech Recognition}

\makeatletter
\def\name#1{\gdef\@name{#1\\}}
\makeatother
\name{{\em Xin Zheng$^{1,2}$, Zhiyong Wu$^{1,2,3}$, Helen Meng$^{1,3}$, Weifeng Li$^{1}$, Lianhong Cai$^{1,2}$}}

\address{$^1$Tsinghua-CUHK Joint Research Center for Media Sciences, Technologies and Systems \\
Shenzhen Key Laboratory of Information Science and Technology\\
Graduate School at Shenzhen, Tsinghua University, Shenzhen 518055, China\\
  $^2$Tsinghua National Laboratory for Information Science and Technology (TNList) \\
  Department of Computer Science and Technology, Tsinghua University, Beijing 100084, China\\
  $^3$Department of Systems Engineering and Engineering Management\\
  The Chinese University of Hong Kong, Shatin, N.T., Hong Kong SAR, China\\
{\small zhengx11@mails.tsinghua.edu.cn, zywu@sz.tsinghua.edu.cn, hmmeng@se.cuhk.edu.hk,}\\
{\small li.weifeng@sz.tsinghua.edu.cn, clh-dcs@tsinghua.edu.cn}}

\begin{document}
\maketitle
\begin{abstract}
In this paper, we first present a new variant of Gaussian restricted Boltzmann machine (GRBM) called multivariate Gaussian restricted Boltzmann machine (MGRBM), with its definition and learning algorithm. Then we propose using a learned GRBM or MGRBM to extract better features for robust speech recognition. Our experiments on Aurora2 show that both GRBM-extracted and MGRBM-extracted feature performs much better than Mel-frequency cepstral coefficient (MFCC) with either HMM-GMM or hybrid HMM-deep neural network (DNN) acoustic model, and MGRBM-extracted feature is slightly better.
\end{abstract}

\section{Introduction}
Since hybrid hidden markov model (HMM)-deep neural network (DNN) was introduced to large-vocabulary continuous speech recognition (LVCSR), the accuracy of speech recognition system has made significant performance improvements in idealized environments \cite{dbnlvcsr}\cite{dnn}. Such progression urges the development of speech recognition systems that are robust to background noise and channel distortion, as more and more speech applications are deploying on mobile devices. State-of-the-art robust automatic speech recognition system usually involves intensive specialized domain knowledge \cite{etsi}. But we are more interested in the transformation of feature.

Feature learning (representation learning) \cite{rl} is a developing field that grows alongside with deep learning. The aim of feature learning is to learn a certain kind of transformation through which we are able to extract information that makes discrimination much easier for classifiers. Feature learning needs as little feature engineering as possible, and transformed feature is much closer to real underlying factors that generate the original features that we observe.

In this paper, we made our first attempt to apply the idea of feature learning to robust speech recognition. Although dozens of alternatives have been proposed over the past few decades, MFCC is still the default choice of feature for many speech applications. Instead of trying to propose another alternative, we are more interested in learning a better representation of MFCC feature with restricted Boltzmann machine (RBM) and its variants. We will explain why RBM may be able to learn a more suitable representation of MFCC for robust speech recognition. We will also be proposing a new variant of RBM called multivariate Gaussian restricted Boltzmann machine (MGRBM) which is specially designed for modeling the distribution of speech data. MGRBM is able to capture the evolving characteristic of speech within a context of several frames which is difficult to model with a Gaussian restricted Boltzmann machine (GRBM). We perform our experiments on the Aurora2 corpus and our results show that the learned features are better than the original feature for robust speech recognition.

\section{Model}
\subsection{Restricted Boltzmann Machine}
The Boltzman machine is a special kind of Markov random field which models the joint probability distribution of the visible variable and hidden variable. Visible and hidden variable are both defined to be multidimensional Bernoulli variables. The distribution can be written as:
\begin{eqnarray}
p(v,h) = \frac{1}{Z}e^{-E(v,h;\theta)}\label{eq:rbm}
\end{eqnarray}
and
\begin{eqnarray}
E(v,h;\theta)=-\frac{1}{2}v^T Uv-\frac{1}{2}h^T Vh-v^T Wh-a^T v-b^T h
\end{eqnarray}
$E(v,h;\theta)$ in (\ref{eq:rbm}) is called energy function. $\theta=\{U,V,W,a,b\}$. $U,V,W$ models the visible-visible, hidden-hidden, and visible-hidden interaction respectively. $a$ and $b$ are bias vectors.

The Restricted Boltzmann Machine (RBM) \cite{rbm} is perhaps the most widely-used variant of Boltzmann machine. The energy function of RBM is the simplified version of that in the Boltzmann machine by making $U=0$ and $V=0$. That is, the energy function of an RBM is:
\begin{eqnarray}
E(v,h;\theta) = -a^{T}v-b^{T}h-v^{T}Wh
\end{eqnarray}

An RBM is typically trained with maximum likelihood estimation. Taking the derivative with respect to the logarithm of the product of all the probability of training cases, we can derive the learning algorithm of RBM as follows:
\begin{eqnarray}
\label{eq:rbmw}\Delta W &=& \epsilon(<vh^{T}>_{data}-<vh^{T}>_{model})\\
\label{eq:rbma}\Delta a &=& \epsilon(<v>_{data}-<v>_{model})\\
\label{eq:rbmb}\Delta b &=& \epsilon(<h>_{data}-<h>_{model})
\end{eqnarray}
The symbol $<\cdot>_{data}$ in (\ref{eq:rbmw})(\ref{eq:rbma})(\ref{eq:rbmb}) represents an average with respect to the conditional distribution $p(v|h)$ and $<\cdot>_{model}$ represents an average with respect to the joint distribution $p(v,h)$. The $<\cdot>_{data}$ for RBM is generally easy to train because:
\begin{eqnarray}
\label{eq:rbmeq1}p(h_j=1|v)=sigmoid(b_j+W_jv)\\
\label{eq:rbmeq2}p(v_i=1|h)=sigmoid(a_i+W_{\cdot i}^Th)
\end{eqnarray}
(\ref{eq:rbmeq1})(\ref{eq:rbmeq2}) can be derived from the definition of RBM and $sigmoid(x)=\frac{1}{1+e^{-x}}$. However, $<\cdot>_{model}$ is much harder to obtain. To address this problem, $<\cdot>_{model}$ is usually approximated with $<\cdot>_{recon}$ as the following:
\begin{eqnarray}
\label{eq:rbmcdw}\Delta W &=& \epsilon(<vh^{T}>_{data}-<vh^{T}>_{recon})\\
\label{eq:rbmcda}\Delta a &=& \epsilon(<v>_{data}-<v>_{recon})\\
\label{eq:rbmcdb}\Delta b &=& \epsilon(<h>_{data}-<h>_{recon})
\end{eqnarray}
The $<\cdot>_{recon}$ represents an average with respect to the reconstruction of the visible data. The reconstruction of visible data is obtained by setting each node in hidden layer value 1 with probability (\ref{eq:rbmeq1}), followed by setting each node in visible layer value 1 with probability (\ref{eq:rbmeq2}). This is the contrastive divergence algorithm (CD) \cite{cd} for training of RBM. CD has been empirically showed to be adequate for many applications.

\subsection{Gaussian Restricted Boltzmann Machine}
\label{sec:grbm}
To model real-valued data, the Gaussian restricted Boltzmann machine (GRBM) has been proposed \cite{reduce}\cite{grbm}. The energy function of GRBM is typically defined with:
\begin{eqnarray}
E(v,h;\theta) = \sum_{i}\frac{(v_{i}-a_{i})^{2}}{2\sigma_{i}^{2}}-\sum_{ij}W_{ij}\frac{v_{i}}{\sigma_{i}}h_{j}-\sum_{j}b_{j}h_{j}
\end{eqnarray}
in which $\theta=\{a,b,W,\sigma\}$ and $\sigma_i$ models the standard deviation of each visible units.

Conveniently, the learning algorithm of GRBM is the same with RBM (\ref{eq:rbmcdw})(\ref{eq:rbmcda})(\ref{eq:rbmcdb}). The conditional probabilities necessary for CD of a GRBM are:
\begin{eqnarray}
\label{eq:grbmeq1}p(v_i|h)=\mathcal{N}(a_i+\sigma_i \sum_j W_{ij} h_j,\sigma_i^2)\\
\label{eq:grbmeq2}p(h_j=1|v)=sigmoid(b_j+\sum_{i}W_{ij}\frac{v_i}{\sigma_i})
\end{eqnarray}

Generally speaking, $\sigma_i$ can be learned from data, but it's difficult with CD. The training data to be modeled with a GRBM are always pre-processed mean 0 and variance 1 and thus $\sigma_i$ can be fixed with 1 and not trained. The reason why $\sigma_i$ is difficult to train with CD can be explained as follows: When $\sigma_i$ is much smaller than 1, the visible-hidden effects (\ref{eq:grbmeq1}) tends to be large and hidden-visible effects (\ref{eq:grbmeq2}) tends to be small. The result of such effect is that hidden units always tend to be firmly 1 or 0, and thus undermine the whole training process.

One disadvantage of GRBM is its conditional independence assumption. That is, conditioned on the hidden layer, each visible unit is assumed to follow a Gaussian distribution and independent with each other. However, for natural data such as speech and image, they tend to have local similarity property. Take speech data for example, the smoothness of speech always makes one frame of acoustic feature similar to the frames next to it. Such local similarity property is difficult to capture with GRBM and yet contains certain amount of information. To offset this problem, we propose a variant of GRBM called multivariate Gaussian restricted Boltzmann machine.


\subsection{Multivariate Gaussian Restricted Boltzmann Machine}
\begin{figure}[htb]
\centerline{\includegraphics[width=8cm]{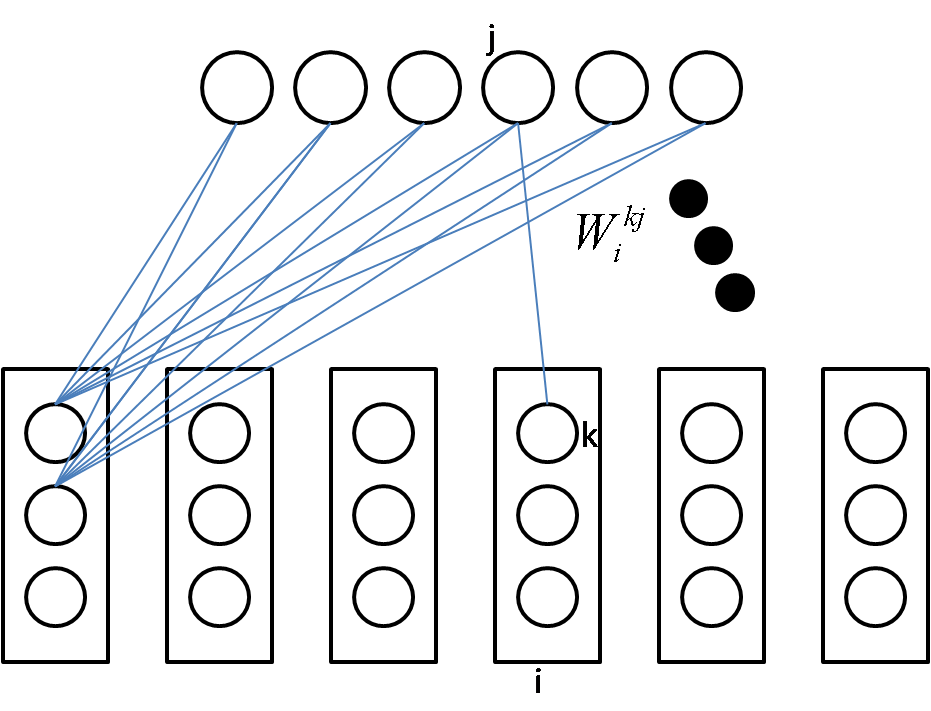}}
\caption{Graphical model of a multivariate Gaussian restricted Boltzmann machine. Links between sub-units in visible layer and hidden units are fully connected. The interaction of the $k$th sub-unit in the $i$th visible unit with $j$th hidden unit is modeled with $W_i^{kj}$.}
\label{fig:mgrbm}
\end{figure}

The multivariate Gaussian restricted Boltzmann machine (MGRBM) is a natural generalization of GRBM. The graphical model of a MGRBM is illustrated in Figure \ref{fig:mgrbm}. Compared with GRBM, in which each unit in visible layer is modeled with a Gaussian distribution given the hidden layer, a MGRBM assumes that each unit in visible layer is modeled with a multivariate Gaussian distribution given the hidden layer. Similar to what $\sigma_i$ models in a GRBM, we denote the covariance matrix of each unit in visible layer of a MGRBM with $\Sigma_i$. Consider only the non-degenerate case, $\Sigma_i$ is a positive-definite matrix and thus Cholesky decomposition can be applied: $\Sigma_i=AA^{T}$. Since matrix $A$ is full-rank, we can denote $B=A^{-1}$. With these notation, we can define the energy function of an MGRBM as:
\begin{eqnarray}
\label{eq:mgrbm}
\begin{split}
E(v,h;\theta)=\frac{1}{2}\sum_{i}(v_{i}-\mu_{i})^{T}B_{i}B_{i}^{T}(v_{i}-\mu_{i})\\
-\sum_{i}v_{i}^{T}B_{i}W_{i}h-b^{T}h
\end{split}
\end{eqnarray}
Suppose the number of units in visible layer and hidden layer is $N_v$ and $N_h$ respectively, and each unit in visible layer has $d$ dimension. Then $v_i,\mu_i$ each is a $d\times 1$ vector; $B_i$ each is a $d\times d$ matrix; $W_i$ each is a $d\times N_h$ matrix; $b$ and $h$ are both $N_h\times 1$ vectors.

Similar to GRBM, we can also prove that:
\begin{eqnarray}
\label{eq:mgrbmeq1}p(v_i|h)=\mathcal{N}(\mu_i+(B_i^T)^{-1}W_i h,(B_i^T)^{-1}B_{i}^{-1})\\
\label{eq:mgrbmeq2}p(h_j=1|v)=sigmoid(b_j+\sum_{i}W_{ij}^T B_i v_i)
\end{eqnarray}
and learning algorithm is:
\begin{eqnarray}
&\begin{split}
\Delta \mu_{i}=\epsilon(<B_{i}B_i^{T}(v_i-\mu_i)>_{data}\\
-<B_{i}B_i^{T}(v_i-\mu_i)>_{model})
\end{split}\\
&\Delta b=\epsilon(<h>_{data}-<h>_{model})\\
&\Delta W_{i}=\epsilon(<B_i^Tv_ih^T>_{data}-<B_i^Tv_ih^T>_{model})\\
&\begin{split}
\Delta B_{i}=\epsilon(<(v_i-\mu_i)(v_i-\mu_i)^TB_i-v_ih^TW_i^T>_{data}\\
-<(v_i-\mu_i)(v_i-\mu_i)^TB_i-v_ih^TW_i^T>_{model})
\end{split}
\end{eqnarray}

In our experiment, we use persistent contrastive divergence (PCD) \cite{pcd} to train MGRBM. The algorithm can be written as:
\begin{eqnarray}
&\begin{split}
\Delta \mu_{i}=\epsilon(<B_{i}B_i^{T}(v_i-\mu_i)>_{data}\\
-<B_{i}B_i^{T}(v_i-\mu_i)>_{fanta})
\end{split}\\
&\Delta b=\epsilon(<h>_{data}-<h>_{fanta})\\
&\Delta W_{i}=\epsilon(<B_i^Tv_ih^T>_{data}-<B_i^Tv_ih^T>_{fanta})\\
&\begin{split}
\Delta B_{i}=\epsilon(<(v_i-\mu_i)(v_i-\mu_i)^TB_i-v_ih^TW_i^T>_{data}\\
-<(v_i-\mu_i)(v_i-\mu_i)^TB_i-v_ih^TW_i^T>_{fanta})
\end{split}
\end{eqnarray}
with $<\cdot>_{fanta}$ denotes average with respect to fantasy particles.

Notice that the problem for updating variances in a GRBM which we described in section \ref{sec:grbm} still exists in MGRBM. In section \ref{sec:rbmtrain} we will explain how we address this problem in our experiments.

\begin{figure}[htb]
\centerline{\includegraphics[width=8.5cm]{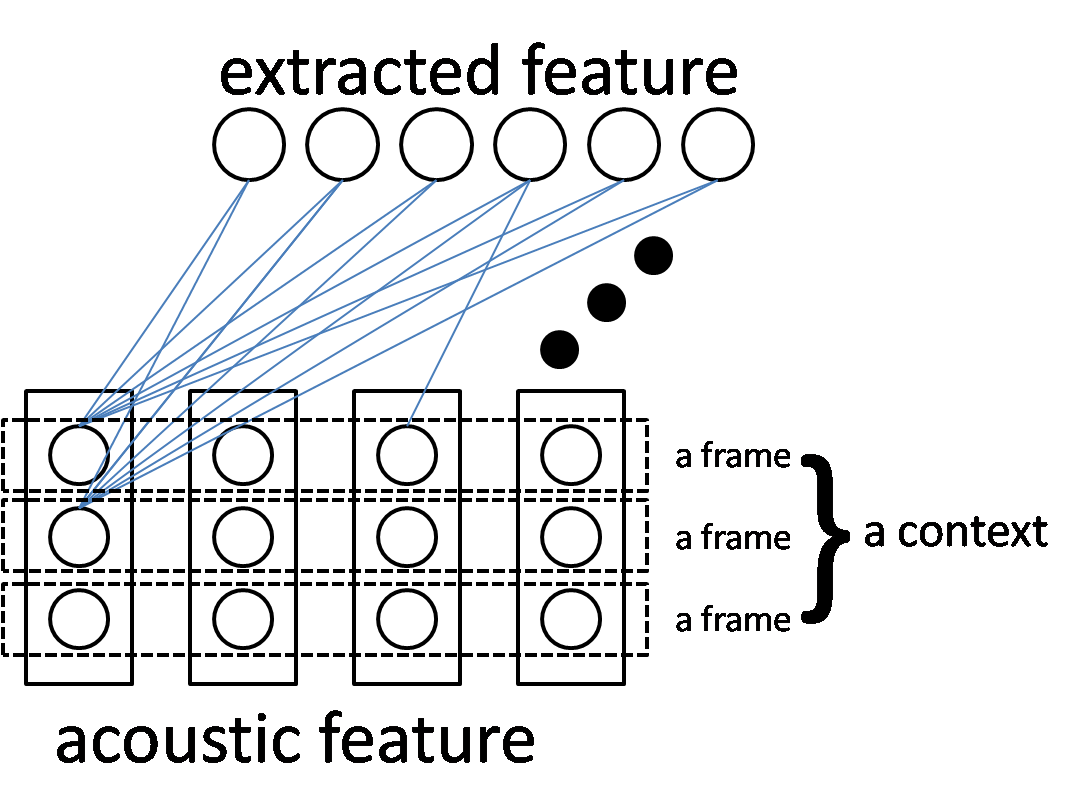}}
\caption{Extract new feature from acoustic feature with multivariate Gaussian restricted Boltzmann machine.}
\label{fig:mgrbmasr}
\end{figure}

MGRBM is specially designed for speech data to address the problem of GRBM described above. Typically, for the task of speech recognition with hybrid HMM-neural network (NN) method, a context of several frames of acoustic feature is used for each training case. Unlike GRBM, MGRBM can explicitly model the evolving characteristics in each context. How a MGRBM can be used for robust feature extraction is illustrated in figure \ref{fig:mgrbmasr}. Concretely, suppose the original acoustic feature for each frame has $D$ dimensions and each context is chosen to be $C$ frames. Then the visible layer of MGRBM has $D$ units, each has dimension $C$; the $d$th dimension in the $c$th frame acoustic feature corresponds to the $c$th dimension of $d$th unit.

There are two reasons for the above setting. First, the correlation modeled across each dimension of acoustic feature would act as a strong regularization in temporal perspective if training data is much different from testing data. Second, as is the case with GRBM, MGRBM also has conditional independence assumption. Fortunately, this assumption is indeed satisfied if we use MFCC as acoustic feature, for the step of discrete cosine transform already has the effect of decorrelation.

\subsection{Feature Learning for Robust Speech Recognition}
\label{sec:fl}
Despite of its prevalence and huge success in phoneme recognition \cite{dbnphone}\cite{dbnacoustic} and LVCSR \cite{dbnlvcsr}, deep neural network (DNN) stand alone is rarely used as a acoustic model for robust speech recognition. We believe that this is due to the fact that neural network is a discriminative model, whose optimization objective is better discriminative power and lower classification error. However, for the task of robust speech recognition, especially in mismatched scenario where the training data is clean and the testing data is noisy, the power of such model is greatly degraded due to its poor generalization over highly distorted data. We assert that DNN performs significantly poorly in very noisy conditions than GMM and we will prove it in our experiments later.

Generative models focus on modeling how the data is ``generated". Natural data such as speech and image are usually high-dimensional, but all that make sense occupy only a subspace (or lower-dimensional manifold). The learning process of a generative model is essentially to find out this manifold by tuning all its parameters. Background noise from real-life environment, we believe, is substantially different from white noise, because the former kind of signal contains certain characteristic shared with sounds that spread through the air. For this reason, we consider it more appropriate to model speech with generative models. 
So, we would like to find a model that can leverage such advantage of generative models and yet escape from a model with strong assumption like GMM, and RBM is indeed a such model.

As a generative model, RBM (and its variants) makes little assumption of data. What is more important, it belongs to a family called product of experts (as opposed to mixture of experts which GMM belongs to) \cite{poe}. This makes it exponentially more powerful and less prone to over-fitting. Other than modeling the distribution of data, it provides a natural way to transform feature by applying (\ref{eq:rbmeq1})(\ref{eq:grbmeq2})(\ref{eq:mgrbmeq2}). Since such transformation takes the whole distribution of data into consideration, the learned feature tends to be much more abstract and expressive.

\section{Experiments}
\label{sec:exp}
In this paper, we used the Aurora2 data set \cite{aurora} for our experiments. In all the experiments described below, we used only clean data set as training data and whole test set as testing data.

We intentionally use acoustic models that are simple and comparable. Two different kinds of acoustic model is utilized in our experiments : HMM-GMM and hybrid HMM-DNN. With each kind of acoustic model, we compare the word error rate (WER) of three kinds of feature : MFCC (12 coefficients + energy + delta + acceleration, 39-dimension), GRBM-extracted feature (G-feature) and MGRBM-extracted feature (M-feature). We used standard Aurora2 setup described in \cite{aurora} for baseline system (GMM + MFCC). 
We use DNN simply because it is a good classifier and it is much more natural to employ RBM-extracted feature.

\subsection{RBM Training}
\label{sec:rbmtrain}
We used two kinds of RBM in our experiment, one for feature extraction and the other for pre-training of DNN.

GRBM and MGRBM are both trained to extract feature for comparison. For these GRBMs and MGRBMs, PCD with stochastic gradient descent (SGD) was used. The size of a mini-batch is 128 and no momentum was applied. The number of fantasy particles was the same with the size of a mini-batch, and one full Gibbs update was performed for each gradient estimate. For all the weights and biases, learning rate was 0.001. For the updating of $B_i$ in (\ref{eq:mgrbm}) of an MGRBM, learning rate was 0.0001. To avoid the problem of updating variances described in section \ref{sec:grbm}, we divide each $B_i$ by its trace after each updating. This step makes all diagonal elements of $B_i$ fixed with one and thus make the learning stable. All GRBM and MGRBM were trained for 400 epochs and each with 1024 hidden units, which made G-feature and M-feature both has 1024 dimensions. The visible layer corresponds to a context of 9 frame of MFCC feature, which makes the number of visible units of GRBM 351 and MGRBM 39$\times$9.

For all the GRBMs and RBMs that were used for pre-training DNNs, CD algorithm (CD-1), SGD with batch size 128 and a momentum of 0.9 were used. We trained all the RBMs with learning rate of 0.01 for 50 epochs and all the GRBMs with learning rate of 0.001 for 100 epochs.

\subsection{DNN Training}
All DNNs in our experiments have 4 hidden layers, each hidden layer with 1024 units. DNNs were pre-trained with stacked RBM (section \ref{sec:rbmtrain}) as described in \cite{dbn} and fine-tuned with back-propagation algorithm with SGD as described in \cite{dbnacoustic}. The learning rate for back-propagation started from 1.0 and was halved if an increase of substitution error on development set was observed during the end of each epoch and all the weights roll-back to the end of last epoch. When using MFCC feature, the input layer of DNN corresponds to 9 frame of MFCC, which is 351 units. 
The output layer has 180 units, with each unit corresponding to each state in HMM. All data to be trained with DNN are normalized to have mean 0 and variance 1 with respect to each dimension.

\subsection{Results}
All results from our experiments are shown in Table 1. We averaged WER in all test set across different noises. The first thing we should notice is that in the lowest SNR condition, HMM-DNN performs consistently poorer than HMM-GMM, which confirms our assertion in section \ref{sec:fl}. Those WER that exceeded 100\% was due to many substitution errors. From the last three columns, we can clearly see the advantage of G-feature over MFCC and M-feature over G-feature. Although the improvements does not seems to be obvious, the trend that the gap between M-feature and G-feature increases with the decrease of SNR is still distinguishable. The reason why the improvements of M-feature over G-feature is marginal, we believe, is that the model which we trained is still not good enough. All the learning rates, number of epochs and number of hidden units are chosen heuristically, and 9 frame of context might be not sufficiently long. So there is no reason to believe we have exploited the full potential of MGRBM.

\begin{table}[h]\footnotesize
\label{tab:ret}
\centering
\vspace{1ex}
\begin{tabular}{c|ccc|ccc}
\hline
\multirow{2}*{\textbf{SNR}}
    &  \multicolumn{3}{c|}{GMM}& \multicolumn{3}{c}{DNN}\\\cline{2-7}

&\thead{\scriptsize{MFCC}} & \thead{\scriptsize{G-feat}\\\scriptsize{(PCA)}} & \thead{\scriptsize{M-feat}\\\scriptsize{(PCA)}} & \thead{\scriptsize{MFCC}} & \thead{\scriptsize{G-feat}} & \thead{\scriptsize{M-feat}} \\
\hline\hline
Clean & 1.05 & 1.94 & 2.75 & 1.27 & 0.76 & 0.82 \\
20dB & 6.10 & 4.28 & 6.21 & 3.76 & 2.93 & 2.49 \\
15dB & 15.18 & 8.50 & 13.67 & 9.09 & 6.47 & 5.64 \\
10dB & 34.82 & 21.64 & 32.06 & 25.85 & 17.37 & 15.79 \\
5dB & 61.27 & 49.80 & 59.75 & 58.05 & 41.87 & 40.99 \\
0dB & 82.56 & 75.15 & 79.89 & 93.16 & 78.32 & 75.28 \\
-5dB & 91.20 & 87.59 & 90.07 & 108.79 & 106.20 & 94.68 \\
\hline
\end{tabular}
\caption{Comparison of different features, all numbers are percentage of WER. G-feat and M-feat here represents G-feature and M-feature respectively.}
\end{table}

With HMM-GMM as acoustic model, the comparison is not so straightforward, because training a 1024-dimensional GMM would leads to severe over-fitting. So we reduced the dimension of G-feature and M-feature to 39 dimensions with principal component analysis (PCA). Notice that this is actually not the right thing to do, because the sum of top 39 eigenvalues only consists 91.8\% and 69.7\% of sum of all eigenvalues for G-feature and M-feature respectively. Despite of such great loss of information, both G-feature and M-feature performs better than MFCC for all SNR levels except the clean speech.

\section{Conclusion and Discussion}
\label{sec:discuss}
In this paper, we briefly reviewed the definition and learning algorithm of RBM and GRBM. We then propose a new variant of RBM called MGRBM by which we would like to model the covariance of adjacent frames within each context. After that we offered an explanation of why a feature learned with RBM (and its variants) might be able to enhance the performance of robust speech recognition over original feature. Finally we performed our experiments on Aurora2 and showed that feature that learned with GRBM and MGRBM would indeed improve the average accuracy across environment in every SNR condition over original MFCC feature.

Throughout the process of feature learning with GRBM, virtually nothing is presupposed. This makes it adaptable with any feature generated from a front-end. Aside from training a GRBM, which can be done off-line, the extra cost of the feature transformation is merely a multiplication with a matrix. From this perspective, feature learning with GRBM is similar to the TANDEM system \cite{tandem}, but the latter is purely supervised. Hence many advantages can be gained if lots of data is accessible but little is labeled. What is more important, GRBM-extracted feature can be used for the TANDEM System seamlessly, which makes TANDEM system semi-supervised and thus further enhancement of performance can be achieved.

\section{Acknowledgements}
This work is partially supported by the National Basic Research Program (973 Program) of China(2012CB316401), the National Natural Science Foundation of China (60928005, 60805008, 60931160443 and 61003094), the Ph.D. Programs Foundation of Ministry of Education of China (200800031015), the Upgrading Plan Project of Shenzhen Key Laboratory and the Science and Technology R\&D Funding of the Shenzhen Municipal.

%
\eightpt

\end{document}